\documentclass[conference]{IEEEtran}
\IEEEoverridecommandlockouts
\newcommand\copyrighttext{%
  \footnotesize \textcopyright 2020 IEEE. Personal use of this material is permitted.  Permission from IEEE must be obtained for all other uses, in any current or future media, including reprinting/republishing this material for advertising or promotional
  purposes, creating new collective works, for resale or redistribution to servers or lists, or reuse of any copyrighted component of this work in other works.}
\newcommand\arxivcopyrightnotice{%
\begin{tikzpicture}[remember picture,overlay]
\node[anchor=south,yshift=10pt] at (current page.south) {\fbox{\parbox{\dimexpr\textwidth-\fboxsep-\fboxrule\relax}{\copyrighttext}}};
\end{tikzpicture}%
}

\usepackage{tikz}
\usepackage{cite}
\usepackage{amsmath,amssymb,amsfonts}
\usepackage{algorithmic}
\usepackage{graphicx,dblfloatfix,caption,afterpage}
\usepackage[font=footnotesize]{caption}
\usepackage{textcomp}
\usepackage{xcolor}
\def\BibTeX{{\rm B\kern-.05em{\sc i\kern-.025em b}\kern-.08em
    T\kern-.1667em\lower.7ex\hbox{E}\kern-.125emX}}
\begin{document}

\title{Learning Options from Demonstration \\ using Skill Segmentation
\thanks{$^\dagger$Authors contributed equally. \newline \newline
\indent This work is based on the research supported in part by the National Research Foundation of South Africa (Grant Number: 17808).}
}

\makeatletter
\newcommand{\linebreakand}{%
  \end{@IEEEauthorhalign}
  \hfill\mbox{}\par
  \mbox{}\hfill\begin{@IEEEauthorhalign}
}
\makeatother

% \author{
%   \IEEEauthorblockN{Matthew Cockcroft}
%   \IEEEauthorblockA{\textit{School of Computer Science and Applied Mathematics} \\
% \textit{University of the Witwatersrand}\\
% Johannesburg, South Africa \\
% Matthew.Cockcroft1@students.wits.ac.za}
% \and
% \IEEEauthorblockN{Shahil Mawjee}
% \IEEEauthorblockA{\textit{School of Computer Science and Applied Mathematics} \\
% \textit{University of the Witwatersrand}\\
% Johannesburg, South Africa \\
% Shahil.Mawjee@students.wits.ac.za}
% \linebreakand
% \IEEEauthorblockN{Steven James}
% \IEEEauthorblockA{\textit{School of Computer Science and Applied Mathematics} \\
% \textit{University of the Witwatersrand}\\
% Johannesburg, South Africa \\
% Steven.James@wits.ac.za}
% \and
% \IEEEauthorblockN{Pravesh Ranchod}
% \IEEEauthorblockA{\textit{School of Computer Science and Applied Mathematics} \\
% \textit{University of the Witwatersrand}\\
% Johannesburg, South Africa \\
% Pravesh.Ranchod@wits.ac.za}
% }

\author{
  \IEEEauthorblockN{Matthew Cockcroft$^\dagger$, Shahil Mawjee$^\dagger$, Steven James, and Pravesh Ranchod}
  \IEEEauthorblockA{\textit{School of Computer Science and Applied Mathematics} \\
\textit{University of the Witwatersrand}\\
Johannesburg, South Africa \\
\{matthew.cockcroft1, shahil.mawjee\}@students.wits.ac.za, \{steven.james, pravesh.ranchod\}@wits.ac.za}
}

\maketitle

\arxivcopyrightnotice

\IEEEpubidadjcol

\begin{abstract}
We present a method for learning options from segmented demonstration trajectories. The trajectories are first segmented into skills using nonparametric Bayesian clustering and a reward function for each segment is then learned using inverse reinforcement learning. From this, a set of inferred trajectories for the demonstration are generated. Option initiation sets and termination conditions are learned from these trajectories using the one-class support vector machine clustering algorithm. We demonstrate our method in the four rooms domain, where an agent is able to autonomously discover usable options from human demonstration. Our results show that these inferred options can then be used to improve learning and planning.
\end{abstract}


\section{Introduction}

Humans often execute tasks in an autonomous manner, without actually focusing on the individual actions they perform. Consider the act of driving a car---the person will enter the car, drive and then arrive at their destination, but while driving they do not contemplate each decision, such as turning, braking or changing gear. Instead, the person only checks whether or not they have arrived at their destination.

Hierarchical reinforcement learning encapsulates this idea for agents acting in an environment. This is achieved through action abstraction, transforming low-level actions into higher-level skills. A skill may be composed of many different actions and consists of a particular endpoint or termination state. Using this approach, an agent executing a specific skill only has to check whether it has reached such a termination state, instead of considering which action to take at each time step.

Before action abstraction can be performed, an agent must first learn which low-level actions to take. One method which has proved successful for teaching agents in complex domains is that of Learning from Demonstration (LfD), which provides the agent with demonstrations created by a human expert from which to learn. The agent then makes use of a process known as inverse reinforcement learning (IRL) \cite{Abbeel2004} to learn a policy that best describes the expert's demonstration.

One issue with this approach is that the learned policy describes the entire demonstration and has no context or transferability outside of the original problem domain. Recent work has attempted to decompose into sets of skills. Most of this work has focused on representing skills as methods which attempt to reach a subgoal state \cite{Konidaris2012}, but these approaches often struggle to encapsulate more complicated reward functions. 

One method which overcomes this is nonparametric Bayesian reward segmentation (NPBRS) \cite{Ranchod2015}. This approach uses nonparametric Bayesian statistics to propose segmentations and maximum entropy IRL \cite{Ziebart2008} to learn the reward functions associated with these segmentations. Another benefit of this method is that it does not require the number of skills to be known, but rather is able to infer them from the data.

While this method has proven to be successful in segmenting skills from demonstrations,
the policies and trajectories generated by NPBRS are not leveraged in any way.
% the main focus is on the reward functions which are obtained, while the policies and trajectories generated by NPBRS are not used in any way.
We propose a method which uses these policies and trajectories so that high-level skills can be autonomously acquired by an agent and leveraged to improve learning.

To transform low-level actions into higher-level skills we make use of the options framework \cite{Sutton1999}. An option is a temporally extended action, so it may take more than one time step to complete. Each option consists of an initiation set, a policy and a termination condition.
To learn options, our method makes use of the one-class support vector machine clustering algorithm \cite{Scholkopf2001}, by classifying the start and endpoints of the trajectories segmented by NPBRS. These classifications provide an initiation set and a termination set, from which a termination condition can be generated. These are  combined with the policies produced by NPBRS to create options.

The method is tested in the four rooms domain used in the original presentation of the options framework \cite{Sutton1999}. We show that an agent in this domain is able to autonomously discover usable options from human demonstration, and is able to use these options to accelerate learning dramatically. We explain the benefits of this method with regards to planning, in that it greatly reduces the size of an agent's decision tree.
% We conclude by addressing some limitations and potential improvements of the method, as well as discussing future applications and implications of this approach.

\section{Background}

\subsection{Reinforcement Learning}

Reinforcement learning defines a set of algorithms which aim to solve problems by maximising the reward obtained in a particular situation. These problems are usually modelled as a Markov Decision Process (MDP)  \cite{Sutton1998}. An MDP is a tuple defined as $M = (S,A,P,R, \gamma)$, where $S$ is a set of states, $A$ is a set of actions, $P(s'|s,a)$ is the probability that a transition to state $s'$ will result from taking action $a$ in state $s$, $R(s,a, s')$ is the reward obtained when taking action $a$ in state $s$ and transitioning to state $s'$, and $\gamma \in [0,1]$  is the discount factor.

The goal of reinforcement learning is to find an action policy that maximises the cumulative expected reward. An action policy is a function which gives the probability of an agent in state $s$ choosing to take action $a$, and is denoted by $\pi(a|s)$. Due to the Markov Property, a transition from one state to the next only depends on the action taken in the current state. 

\subsection{Options Framework}

While any task can theoretically be solved using only low-level actions, in practice many learning tasks are intractable in this space. Action abstraction, which combines low-level actions into higher-level skills, can be used to reduce the learning time. A commonly-used approach for this is the options framework \cite{Sutton1999}. An option is a temporally-extended action made up of the original low-level actions. 

An option $o$ is formally defined as a tuple $(I_o,\pi_o,\beta_o)$, where $I_o \subseteq S$ is an initiation set of all states from which option $o$ can be initiated, $\pi_o:  S \times A \rightarrow [0,1]$ is a policy giving the probability of option $o$ executing each action in each state, and $\beta_o:  S \rightarrow [0,1]$ is the termination condition, which gives the probability of the option terminating in a particular state.

A method for creating a new option needs to determine how to learn its policy and how to define its initiation set and termination condition. This is achieved by identifying states in which the option will terminate once reached, and by defining the initiation set as the set of states from which that option would be useful to execute.

\subsection{Inverse Reinforcement Learning}

Inverse reinforcement learning (IRL) \cite{Abbeel2004} defines a set of algorithms used to determine a reward function from a given demonstration. IRL algorithms consider all of the information from an MDP excluding the reward function, defined as an MDP$\backslash R$. Together with this they also require a set of demonstration trajectories, defined as $\zeta=\{(s_1, a_1),(s_2, a_2),...,(s_n, a_n)\}$, where each pair $(s_i,a_i)$ indicates the action $a_i$ taken by the demonstrator while in state $s_i$. The algorithm then attempts to find the reward function $R$ that is most likely to have produced $\zeta$ by an agent attempting to solve the given MDP $(S,A,P,R, \gamma)$.

We consider the maximum entropy inverse reinforcement learning algorithm \cite{Ziebart2008}. This algorithm builds upon the initial approach to IRL \cite{Abbeel2004}, which maps rewards to features within the states so as to reflect the importance of those features to the expert presenting the demonstration. The problem is that IRL is ill-posed, as it is possible for many reward functions to map to the same feature counts.

Maximum entropy IRL instead focuses on the entire distribution of possible behaviours and considers a set of trajectories, $\zeta$. Using MDPs as previously defined, maximum entropy IRL additionally defines $\mathbf{f_s} \in \mathbf{R^k}$ as the feature vector of the state $s$, and $\theta \in \mathbf{R^k}$ as the reward weights \cite{Ziebart2008}. %The reward of a trajectory is then given by 
%$$R(\zeta) = \sum_{s \in \zeta} \theta^T \mathbf{f_s}.$$

The algorithm uses the principle of maximum entropy \cite{Jaynes1957} to assert that the most likely distribution is one that does not display any preferences not implied by the feature counts. This is achieved by the relation $P(\zeta) \propto e^{R(\zeta)}$, where $P(\zeta)$ is the probability of trajectory $\zeta$ occurring. In this model, paths with higher rewards are exponentially preferred over paths with lower rewards. This is used to generate a distribution over paths, given by $P(\zeta|\theta,P)$, where $P$ is the transition distribution. By maximising the entropy of the distribution over paths a convex function is obtained, for which its maxima can be calculated using gradient-based optimization. %This gradient is the difference between expected empirical feature counts, $\Tilde{f}$, and the sum of the learner's expected feature counts, $D_{s_i} f_{s_i}$, where $D_{s_i}$ is the expected state visitation frequencies. The gradient is then defined as 
%$$\nabla L(\theta) = \Tilde{f}-\sum_{s_i} D_{s_i} f_{s_i}.$$

At the maxima, the gradient is $0$, and so the feature counts match. This guarantees that the agent performs equivalently to the expert's demonstration.

\subsection{Nonparametric Bayesian Reward Segmentation}

Bayesian statistical inference is the process of calculating a posterior distribution by using a given prior probability distribution for an unknown parameter and a calculated likelihood for that distribution. Nonparametric models \cite{Hollander2013} are probability models that may have infinitely many parameters, while stochastic processes, such as the Beta process, are examples of prior distributions when dealing with a nonparametric Bayesian model. 

Nonparametric Bayesian reward segmentation (NPBRS) \cite{Ranchod2015} is a method proposed for combining nonparametric Bayesian statistics for segmentation and IRL for policy learning. The maximum entropy algorithm is used to find the option policies that represent skills and the Beta-Process Autoregressive Hidden Markov Model (BP-AR-HMM) \cite{Fox2014} is used to determine how to segment the demonstration for skill extraction.

A Hidden Markov Model is an MDP with hidden states, known as modes. The benefit of using the BP-AR-HMM is that the number of hidden modes does not need to be known as a beta process prior is placed on the sequence of modes. Instead of specifying the appropriate number of modes, they can then be inferred directly from the data. The model is also autoregressive, meaning that for continuous observations a mode-specific Vector Autoregressive (VAR) process can be used to describe temporal dependencies. The generative model for the BP-AR-HMM \cite{Fox2014} is given as follows:

\begin{align}
	B\mid B_0 &\sim BP(1,\beta) \nonumber\\
    X_i\mid B &\sim BeP(B) \nonumber\\
    \pi_j^{(i)}|f_i, \gamma, \kappa &\sim Dir([\gamma,...,\gamma + \kappa,\gamma,...]\otimes f_i) \nonumber\\
    z_t^{(i)} &\sim \pi_{z_{t-1}^{(i)}}^{(i)} \nonumber\\
    y_t^{(i)} &= \sum_{j=1}^{r} A_{j,z_t^{(i)}} y_{t-j}^{(i)} + e_t^{(i)}(z_t^{(i)}) \nonumber
\end{align}

$B$ is drawn from a Beta Process (BP) and provides a set of global probabilities for each skill. This probabilities are used to produce a Bernoulli Process (BeP), from which $X_i$ is drawn. Considering the $ith$ trajectory, distribution $X_i$ is used to construct a binary feature vector $f_i$, indicating which skills are present for this trajectory. Note that $B$ encourages sharing of skills among the demonstration trajectories. Next, for mode $j$ we define the transition probabilities for time series $i$ using the transition probability vector $\pi_j^{(i)}$, which is drawn from a Dirichlet distribution.

Hyperparameter $\kappa$ is used to place extra expected mass on the $jth$ component, making the selection ``sticky" due to the fact that skills are likely to be employed for multiple sequential time steps. For each time step $t$, a mode $z_t^{(i)}$ is drawn from the transition distribution at time step $t-1$. The observation for the $ith$ time series, at time $t$, is represented by $y_t^{(i)}$. If the order of the model is $r$, then $y_t^{(i)}$ is computed as a sum of mode-dependent linear transformations using the previous $r$ observations, as well as the model-dependent noise term $e_t^{(i)}$. 

Sampling of the mode sequence $z_t^{(i)}$ is achieved using the Markov Chain Monte Carlo sampler \cite{Fox2009} developed for the BP-AR-HMM. This sampler proposes skill birth and death moves based on their likelihoods, thereby adding or removing features from the global feature set.

The BP-AR-HMM emissions are modelled as VAR processes by \textit{Fox et al.} \cite{Fox2014}. To use this model in conjunction with inverse reinforcement learning, we require the emissions to be modelled as MDPs. \textit{Ranchod et al.} \cite{Ranchod2015} propose the following model to achieve this: 
\begin{align}
	P(a)|z_t^{(i)} &= \frac{e^{\tau Q^{z_t^{(i)}}(y_{t-1}^{(i)},a)}}{\sum_{a} e^{\tau Q^{z_t^{(i)}}(y_{t-1}^{(i)},a)}} \nonumber\\
    a_t^{(i)} &\sim P(a)|z_t^{(i)} \nonumber\\
    y_t &\sim T(y_{t-1},a_t^{(i)}) \nonumber
\end{align}
This model removes the step of sampling of parameters $A$ and $e$ for the BP-AR-HMM. Instead the dynamics of the environment are used in conjunction with IRL to calculate the transition probabilities. The action-value function associated with each skill is represented by $Q^{z_t^{(i)}}$. This function is learned by grouping sub-trajectories in terms of skills and performing the maximum entropy IRL algorithm on each skill. This produces reward functions associated with each skill. The optimal policy for $Q^{z_t^{(i)}}$ is then determined by using value iteration, which gives the likelihood of each of the demonstration trajectories, and selecting the set with the highest likelihood \cite{Ranchod2015}.

\section{Research Methodology}

While previous work using the NPBRS framework \cite{Ranchod2015} is able to successfully segment skills from demonstrations, the resulting policies are not used in any way. In this section, we show how to leverage these approaches to learn option models, consisting of initiation sets, policies and termination conditions. This requires creating a set of demonstration trajectories, segmenting these trajectories using NPBRS, and then estimating the initialisation set and termination condition from the segmented trajectories produced by NPBRS. 

\subsection{Generating Trajectories}

Due to the deterministic nature of the domain in which this approach was tested, the use of human expert demonstrations was not required. Instead, Q-learning \cite{Watkins1992} was used to learn an optimal policy for the domain and trajectories were generated by following this policy to and from predefined start and endpoints. These trajectories are then sent to the NPBRS method to be segmented.

\subsection{Learning Options from Demonstration}

Given the set of trajectories produced by NPBRS, we wish to generate estimates for the initialisation and termination sets for an option. We phrase this as a classification problem in which we wish to classify the final trajectory state as a termination state (positive) or a non-termination state (negative), and similarly classify start states as to whether they are initiation states or not. It is important to note here that while we are classifying termination states, options require a termination condition. Thus, we will use our set of termination states to generate a probability of being a termination state for any particular state and use this as our termination condition.

A commonly used algorithm for supervised classification problems is the Support Vector Machine (SVM) \cite{Vapnik2013}. The SVM aims to split a given dataset into two separate classes by finding a hyperplane with the maximum margin between the two classes. SVMs require the training data to be labelled into two distinct classes, but in our case there are no negative samples available for learning the initiation set and termination condition of the options. 

An alternate approach is to use an anomaly detection method, in this case the one-class SVM (OC-SVM) \cite{Scholkopf2001}. The OC-SVM determines a hypersphere that bounds as much of the training data as possible, while attempting to minimise its volume.

Given a set of training data ${X_1, ..., X_n} \in d$, the OC-SVM first uses a mapping $\Phi: \mathcal{R}^{d} \rightarrow \mathcal{F}$  to project this data into a higher-dimensional space. The hypersphere in this space is then parameterised by a centre $c$ and a radius $r$. These are computed by minimising the equation:

$$
\begin{array}{c}{\min\limits_{r, c, \xi} r^{2}+\frac{1}{v n} \sum_{i} \xi_{i}} \\ \\
{\text { s.t. }\left\|\Phi\left(X_{i}\right)-c\right\|^{2} \leq r^{2}+\xi_{i}, \quad \xi_{i} \geq0, \quad i=1, \ldots, n},
\end{array}
$$

where $v \in (0, 1)$ is a trade-off parameter between the radius and the number of training data points that fall inside the hypersphere, and $\xi_{i}$ are slack variables which determine how far away outliers lie from the hypersphere surface.

Thus, given a set of trajectories, we can use the OC-SVM to define separate hyperspheres, using the trajectory start and end states as our training data. We will then consider any state which lies within our hypersphere trained on the start states as an initiation state and any state which lies within our hypersphere trained on the end states as a termination state.

\section{Experimentation}

\subsection{Domain and Setup}

The domain used to test our method for learning options is the four rooms domain discussed in the original options framework paper \cite{Sutton1999}. The domain consists of four rooms, each connected by a hallway. \textit{Sutton et al.} \cite{Sutton1999} investigates the usefulness of options by handcrafting options for each room that take the agent to the adjoining hallways. In this case the initiation set is any state within the room, the policy is the set of actions taking the shortest path to the hallway and the termination condition is 1 for a hallway state and 0 elsewhere. 

For each room, there are two different options available, one which takes the agent from its current position to the hallway encountered when travelling clockwise, and the other which takes the agent to the corresponding anti-clockwise hallway. We wish to infer similar options from a set of given demonstrations.

% \begin{figure}[b!]
%   \centering
%   \includegraphics[width=0.25\textwidth]{Images/Four_Rooms.png}
%   \caption{Four rooms domain, with hallways highlighted in green}
%   \label{fig:hallways}
% \end{figure}

First, we use Q-learning to generate a set of 5000 trajectories, each with random start and goal states. The reward function for learning in this domain was defined as 10 at the goal state and -1 elsewhere. These trajectories were then used as input for the NPBRS segmentation, with the sampler left to run for 60 minutes.
This produced a total of four segmented skills, each with their own policy, reward function and set of inferred trajectories. 
%Using the obtained reward functions we can see that NPBRS does well at detecting hallway areas, as seen in Fig. \ref{fig:rewardfunc}, with a threshold of 0.12 applied to the reward values.

% \begin{figure}[b!]
%   \centering
%   \includegraphics[width=0.25\textwidth]{Images/Reward_function_2.png}
%   \caption{Thresholded Reward function for a particular skill}
%   \label{fig:rewardfunc}
% \end{figure}

Due to the fact that a large number of trajectories were used to generate our skills, the number of inferred trajectories returned by NPBRS for each skill is also large. We wish to only use trajectories in which the termination state has a high likelihood of occurring. To achieve this we apply a threshold of 2\% of the total states on the occurrence of termination states and discard any trajectory with a termination state occurring less than this threshold.

Using the trajectories resulting from this thresholding we can then apply our OC-SVM to obtain sets of initiation and termination states. We use the \textit{Scikit-learn} one-class SVM for Python, with the radial basis function kernel, $\nu = 0.1$ and $\gamma = 0.5$, where $\nu$ is an upper bound on the fraction of training errors and $\gamma$ is the kernel coefficient.

\subsection{Results and Discussion}

The set of initiation states obtained from applying the OC-SVM to the trajectory sets can be seen in Fig. \ref{fig:initiations} and similarly the set termination states can be seen in Fig. \ref{fig:terminations}. We define our termination condition here as 1 for any state contained in the set of termination states and 0 elsewhere. Finally, to obtain the policies of our options, we take the policy generated for each skill by NPBRS and restrict it to only include states that the policy leads to from any of our initiation states. These policies can be seen in Fig. \ref{fig:policies}, where green states indicate our initiation states and purple states indicate termination states. 

%\afterpage{\newpage}

\begin{figure*}[!ht]
\centering
    \includegraphics[width=0.20\textwidth]{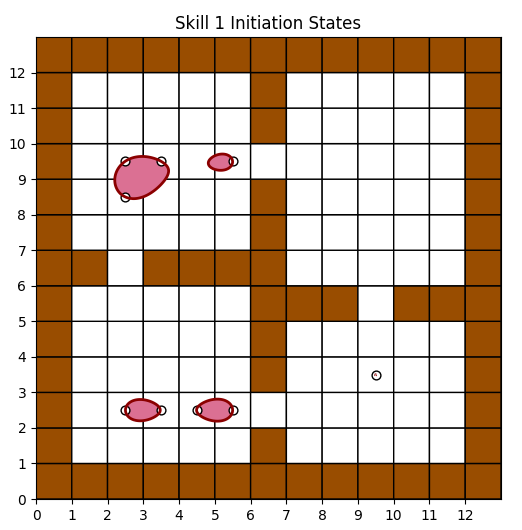}\hfil
    \includegraphics[width=0.20\textwidth]{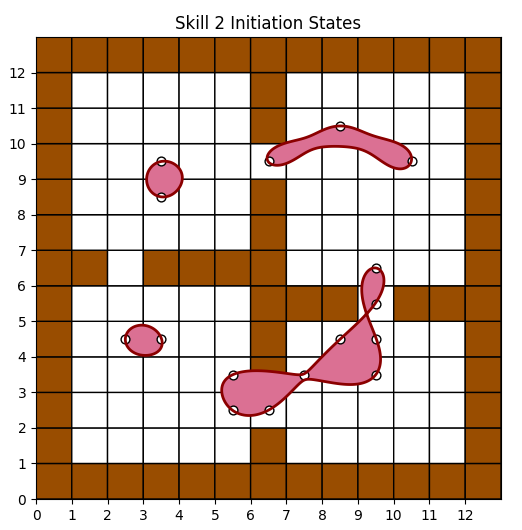}\hfil
    \includegraphics[width=0.20\textwidth]{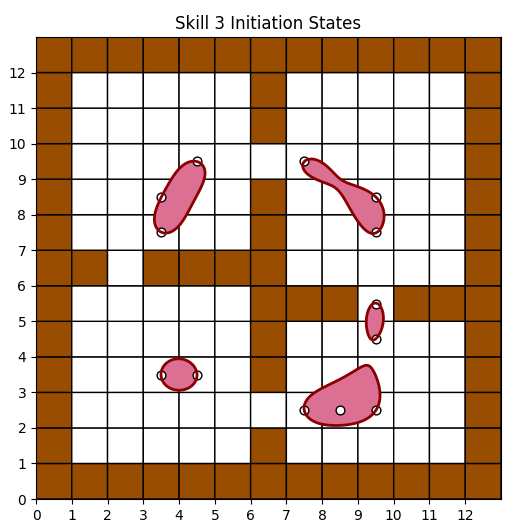}\hfil
    \includegraphics[width=0.20\textwidth]{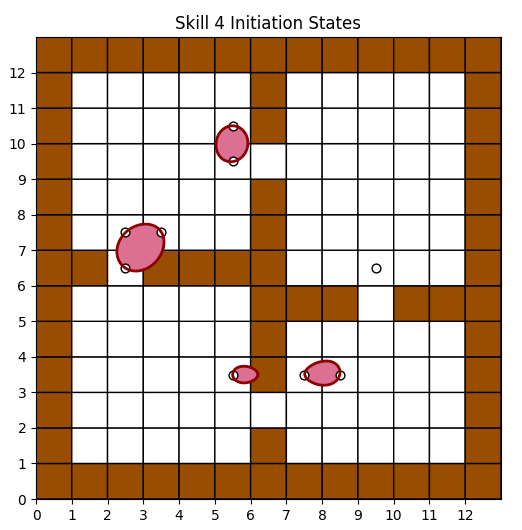}
\caption{Initiation states learned by OC-SVM}
\label{fig:initiations}
\end{figure*}

%\afterpage{\newpage} 

\begin{figure*}[!ht]
\centering
    \includegraphics[width=0.21\textwidth]{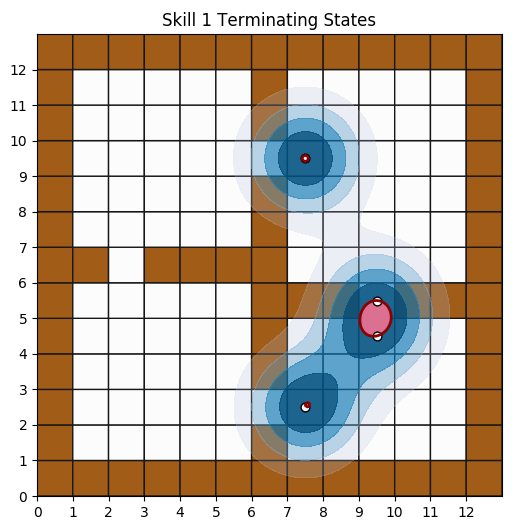}\hfil
    \includegraphics[width=0.21\textwidth]{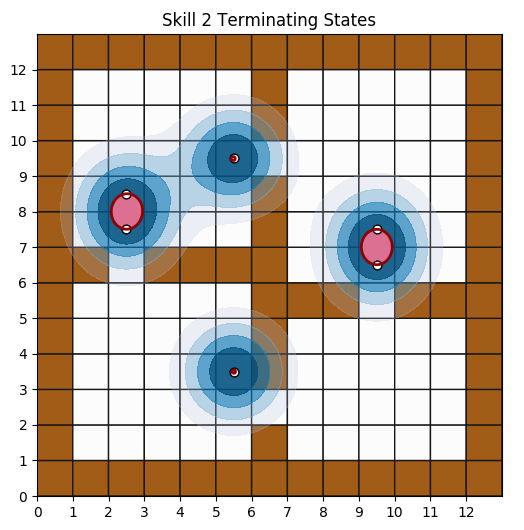}\hfil
    \includegraphics[width=0.21\textwidth]{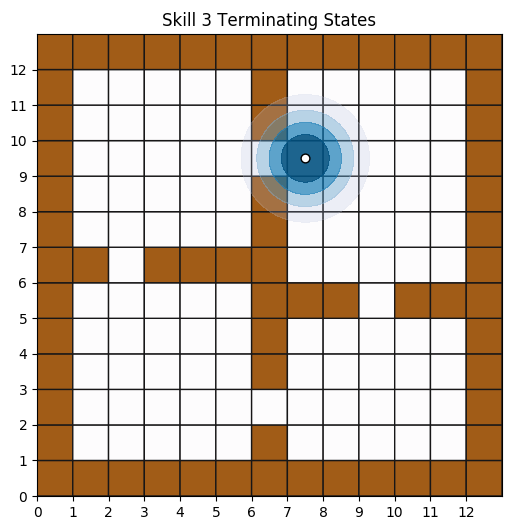}\hfil
    \includegraphics[width=0.21\textwidth]{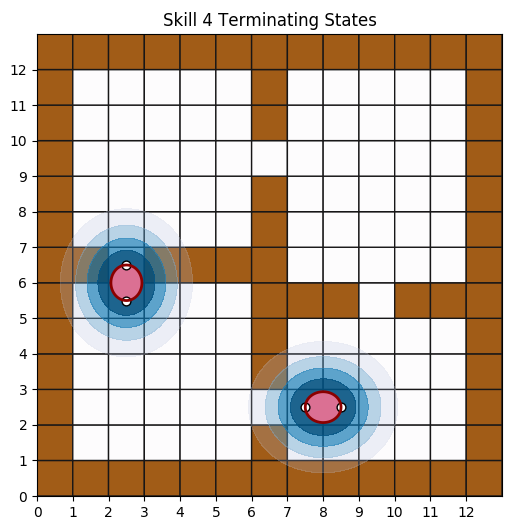}
\caption{Termination states learned by OC-SVM}
\label{fig:terminations}
\end{figure*}

%\afterpage{\newpage} 

\begin{figure*}[!ht]
\centering
    \includegraphics[width=0.21\textwidth]{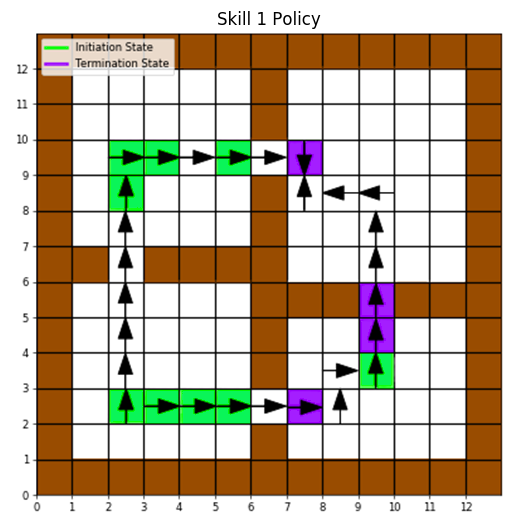}\hfil
    \includegraphics[width=0.21\textwidth]{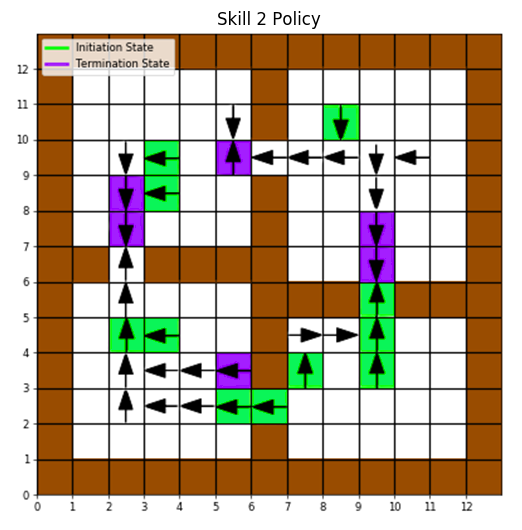}\hfil
    \includegraphics[width=0.21\textwidth]{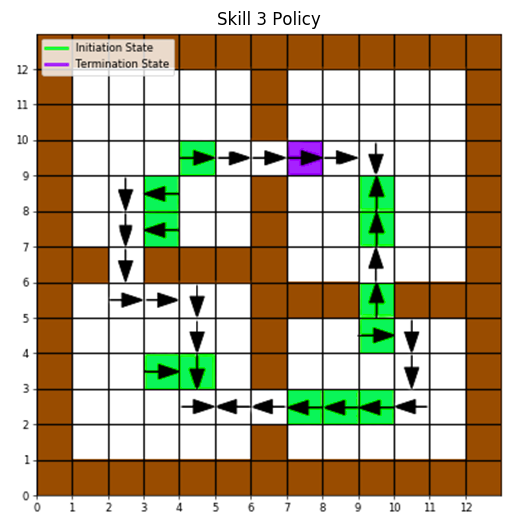}\hfil
    \includegraphics[width=0.21\textwidth]{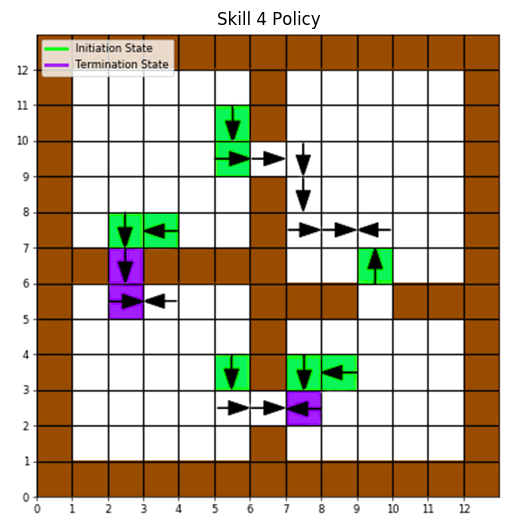}
\caption{Policies for each skill starting at initiation states}
\label{fig:policies}
\end{figure*}

% \afterpage{\newpage} 

Looking at the resulting options that we have obtained, we put particular focus on the termination condition. This is because it is these termination conditions which assist in greatly reducing the complexity of the agent's decision tree. Based on how we have defined our termination condition, the agent no longer has to check which action to take at each time step once initiating an option. Instead it is only required to check whether it is in a termination state or not. 

From the termination states detected in Fig. \ref{fig:terminations} we observe that the OC-SVM identifies hallway areas as the termination states. We consider the three states surrounding a particular hallway as states contributing to a \textit{hallway zone}. This is because these states act as chokepoints for any trajectory moving between rooms, as the agent will have to pass through all three of these states. All of the termination states that we obtain are within 1 step of these \textit{hallway zone} states.

It is easiest to consider the initiation states in Fig. \ref{fig:initiations} in conjunction with the policies beginning from those states in Fig. \ref{fig:policies}. From this we can see that the initiation states typically lie on a path that takes the agent between hallways. Hence the options that we form by combining our policy, initiation set and termination condition allow for the agent to travel between hallways. The options that we are able to autonomously detect are very similar to those defined originally for the four rooms domain by \textit{Sutton et al.} \cite{Sutton1999}. 

% The options we wished to obtain would take the agent to either the clockwise or the anti-clockwise hallways for each room. This can be achieved by switching between the set of options that we have learned. For example, using the policies shown in Fig. 3, starting in the top right room and following each policy to a termination state before switching to a new policy, we can travel to every hallway area in a clockwise direction by switching between policies 1, 4, 3 and 2. Similarly we can obtain an option to take us to the anti-clockwise hallways by switching between skills policies 2, 3 and 1.

The only notable significance between the sets of options is that the original initiation set contains all states for a particular room, while our initiation set only lies on the path between hallways, restricting us to inter-hallway travel. This aside, we can conclude that our method was successful in learning options from demonstration trajectories for the four rooms domain.

\subsection{Learning With Options}

To show the effectiveness of our learned options, we use them to find optimal policies for two different goal states. This process mimics the learning performed in the original options work \cite{Sutton1999}, with the goal states defined as G1 and G2 as seen in Fig. \ref{fig:fourrooms}.

\begin{figure}[b!]
  \centering
  \includegraphics[width=0.21\textwidth]{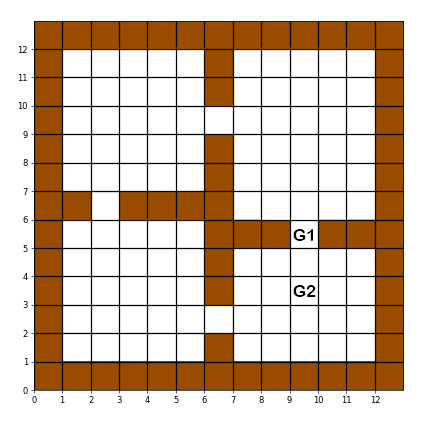}
  \caption{Four rooms domain with goals defined at G1 and G2}
  \label{fig:fourrooms}
\end{figure}

We combine our options with the standard MDP for the four rooms domain to form a semi-Markov decision processes (SMDP) \cite{bradtke1995}, which allows us to switch between executing option and using single actions.

We apply Q-learning \cite{Watkins1992} to our domain to learn the optimal policies, recording the average number of steps taken for each episode. We then compare this to the average number of steps using Q-Learning without options, and using the handcrafted clockwise and anti-clockwise options described previously. For all cases, Q-learning is performed with $\epsilon$-greedy exploration ($\epsilon = 0.1$), learning rate $\alpha=0.5$, and discount factor $\gamma = 0.9$. Results are averaged over 25 runs.

The results are shown in Fig. \ref{fig:graphs}, where our options display similar performance to that of the handcrafted options. Both SMDPs begin learning at a lower average number of steps and converge faster than the MDP without options.

\begin{figure*}[!ht]
\centering
    \includegraphics[width=0.3\textwidth]{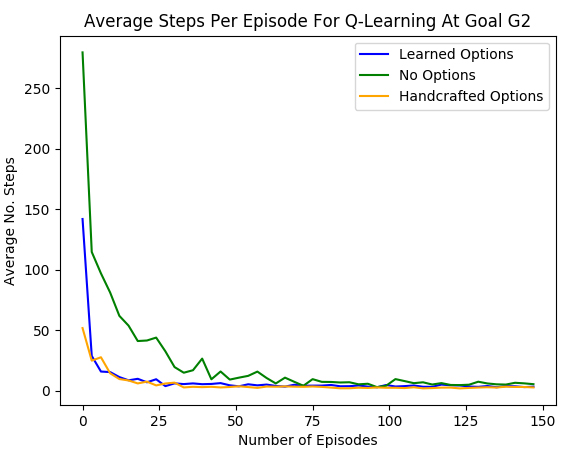}\hfil
    \includegraphics[width=0.3\textwidth]{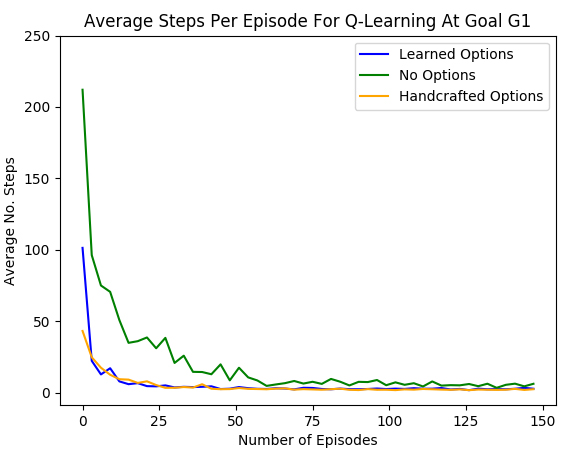}
\caption{Performance comparison between the standard MDP without options, our learned options, and the handcrafted options from the original framework using Q-learning in the four rooms domain.}
\label{fig:graphs}
\end{figure*}

The difference between our options and the handcrafted options can be seen in the first few episodes, where our options have a higher average number of steps due to the limited initiation sets in comparison to the handcrafted initiation sets which each consist of the entire room.

\subsection{Limitations and Future Work}

One of the main limitations with using the OC-SVM is in generating the termination condition. This is due to the fact that the OC-SVM is used for anomaly detection and thus only returns a classification of -1 or 1 depending on whether a given point is predicted to be an anomaly or not. This does not prove to be an issue in our study because we define our termination condition to be 1 for a state in the set of termination states and 0 elsewhere, but generally we want to be able to estimate the probability of terminating at each state. Potential adaptations to achieve this include running the output through logistic regression or using a kernel density estimator \cite{Wand1994}. 

Another limitation is in the number of trajectories we have generated. NBPRS should generally perform better given more trajectories, but it is possible that too many trajectories can result in noise occurring in the skills that are proposed. Ideally we would like to generate multiple sets of trajectories of different sizes and use the set for which the segmentation of the demonstration has highest log-likelihood. %This was not possible in this case though due to time constraints.

Finally, we must note that the NPBRS framework looks to find the longest trajectory possible to explain a single skill. In this case the domain is not very large and so smaller trajectories might explain the skill better. Despite this, both of the estimated policies will still be the same and so this should not have too great of an impact on the final options that are learned.

One potential extension of the work in this area would be to incorporate the learned termination states into the NPBRS framework. These termination states can then be used to cap the length of the proposed trajectories. Another potential area of future work is the adaptation of this method to a continuous domain, and further into real-world domains for robot tasks.

\section{Related Work}

\textit{Niekum et al.} \cite{Niekum2012} also make use of the BP-AR-HMM for segmenting demonstration trajectories, but model the extracted skill policies as Dynamic Movement Primitives rather than options. The use of a sequence of primitives allows for better generalisation and the application of policy improvement on the skill policies, but is limited to single demonstration segments. 

Support vector machines have been used in conjunction with the options framework before to learn sets of parameterised skills \cite{DaSilva2012}. This method requires the policies for these skills to be  learned from experience rather than segmented from a set demonstration trajectories though.

An alternate approach to learning options from sets of demonstration trajectories has been through the use of clustering algorithms. This includes K-means clustering and spectral clustering methods such as Perron Cluster Analysis \cite{Ng2002,Lakshminarayanan2016}. These methods group regions of the trajectory into abstract states and options are then used to define transitions between these states.

%There has also been recent work done in presenting IRL methods which improve upon the performance of the maximum entropy IRL algorithm, such as Generative Adversarial Imitation Learning \cite{Ho2016}. The NPBRS framework remains open to improvement through the adoption of such algorithms over the existing use of maximum entropy IRL.

\section{Conclusion}

In this paper, we presented a method that allows an agent to autonomously discover usable options from human demonstrations. We detailed the method of discovering these skills using the one-class SVM and tested it the four rooms domain, showing the benefit that the discovered options provide for learning. Finally, we outlined areas of improvement within the method and discussed the potential to incorporate the method within the NPBRS framework. While the method has proved successful in a common reinforcement learning domain, there is also potential for it to be adapted to handling more complex, real-world problems.

\bibliographystyle{IEEEannot}
\bibliography{main}

\begin{thebibliography}{10}
\providecommand{\url}[1]{#1}
\csname url@rmstyle\endcsname
\providecommand{\newblock}{\relax}
\providecommand{\bibinfo}[2]{#2}
\providecommand\BIBentrySTDinterwordspacing{\spaceskip=0pt\relax}
\providecommand\BIBentryALTinterwordstretchfactor{4}
\providecommand\BIBentryALTinterwordspacing{\spaceskip=\fontdimen2\font plus
\BIBentryALTinterwordstretchfactor\fontdimen3\font minus
  \fontdimen4\font\relax}
\providecommand\BIBforeignlanguage[2]{{%
\expandafter\ifx\csname l@#1\endcsname\relax
\typeout{** WARNING: IEEEtran.bst: No hyphenation pattern has been}%
\typeout{** loaded for the language `#1'. Using the pattern for}%
\typeout{** the default language instead.}%
\else
\language=\csname l@#1\endcsname
\fi
#2}}

\bibitem{Abbeel2004}
P.~Abbeel and A.~Y. Ng, ``Apprenticeship learning via inverse reinforcement
  learning,'' in \emph{Proceedings of the twenty-first international conference
  on Machine learning}.\hskip 1em plus 0.5em minus 0.4em\relax ACM, 2004, p.~1.


\bibitem{bradtke1995}
S.~J. Bradtke and M.~O. Duff, ``Reinforcement learning methods for
  continuous-time markov decision problems,'' in \emph{Advances in neural
  information processing systems}, 1995, pp. 393--400.


\bibitem{DaSilva2012}
B.~C. Da~Silva, G.~Konidaris, and A.~G. Barto, ``Learning parameterized
  skills,'' in \emph{Proceedings of the 29th International Coference on
  International Conference on Machine Learning}.\hskip 1em plus 0.5em minus
  0.4em\relax Omnipress, 2012, pp. 1443--1450.


\bibitem{Fox2014}
E.~B. Fox, E.~B. Sudderth, M.~I. Jordan, and A.~S. Willsky, ``Joint modeling of
  multiple related time series via the beta process,'' \emph{The Annals of
  Applied Statistics}, vol.~8, no.~3, pp. 1281--–1313, 2014.


\bibitem{Fox2009}
E.~B. Fox, ``Bayesian nonparametric learning of complex dynamical phenomena,''
  Ph.D. dissertation, Massachusetts Institute of Technology, 2009.


\bibitem{Hollander2013}
M.~Hollander, D.~A. Wolfe, and E.~Chicken, \emph{Nonparametric statistical
  methods}.\hskip 1em plus 0.5em minus 0.4em\relax John Wiley \& Sons, 2013,
  vol. 751.


\bibitem{Jaynes1957}
E.~T. Jaynes, ``Information theory and statistical mechanics,'' \emph{Physical
  review}, vol. 106, no.~4, p. 620, 1957.


\bibitem{Konidaris2012}
G.~Konidaris, S.~Kuindersma, R.~Grupen, and A.~Barto, ``Robot learning from
  demonstration by constructing skill trees,'' \emph{The International Journal
  of Robotics Research}, vol.~31, no.~3, pp. 360--375, 2012.


\bibitem{Lakshminarayanan2016}
A.~S. Lakshminarayanan, R.~Krishnamurthy, P.~Kumar, and B.~Ravindran, ``Option
  discovery in hierarchical reinforcement learning using spatio-temporal
  clustering,'' \emph{arXiv preprint arXiv:1605.05359}, 2016.


\bibitem{Ng2002}
A.~Y. Ng, M.~I. Jordan, and Y.~Weiss, ``On spectral clustering: Analysis and an
  algorithm,'' in \emph{Advances in neural information processing systems},
  2002, pp. 849--856.


\bibitem{Niekum2012}
S.~{Niekum}, S.~{Osentoski}, G.~{Konidaris}, and A.~G. {Barto}, ``Learning and
  generalization of complex tasks from unstructured demonstrations,'' in
  \emph{2012 IEEE/RSJ International Conference on Intelligent Robots and
  Systems}, Oct 2012, pp. 5239--5246.


\bibitem{Ranchod2015}
P.~{Ranchod}, B.~{Rosman}, and G.~{Konidaris}, ``Nonparametric bayesian reward
  segmentation for skill discovery using inverse reinforcement learning,'' in
  \emph{2015 IEEE/RSJ International Conference on Intelligent Robots and
  Systems (IROS)}, Sep. 2015, pp. 471--477.


\bibitem{Scholkopf2001}
B.~Sch{\"o}lkopf, J.~C. Platt, J.~Shawe-Taylor, A.~J. Smola, and R.~C.
  Williamson, ``Estimating the support of a high-dimensional distribution,''
  \emph{Neural computation}, vol.~13, no.~7, pp. 1443--1471, 2001.


\bibitem{Sutton1998}
R.~S. Sutton, A.~G. Barto, \emph{et~al.}, \emph{Introduction to reinforcement
  learning}.\hskip 1em plus 0.5em minus 0.4em\relax MIT press Cambridge, 1998,
  vol. 135.


\bibitem{Sutton1999}
R.~S. Sutton, D.~Precup, and S.~Singh, ``Between mdps and semi-mdps: A
  framework for temporal abstraction in reinforcement learning,''
  \emph{Artificial intelligence}, vol. 112, no. 1-2, pp. 181--211, 1999.


\bibitem{Vapnik2013}
V.~Vapnik, \emph{The nature of statistical learning theory}.\hskip 1em plus
  0.5em minus 0.4em\relax Springer science \& business media, 2013.


\bibitem{Wand1994}
M.~P. Wand and M.~C. Jones, \emph{Kernel smoothing}.\hskip 1em plus 0.5em minus
  0.4em\relax Chapman and Hall/CRC, 1994.


\bibitem{Watkins1992}
C.~J. Watkins and P.~Dayan, ``Q-learning,'' \emph{Machine learning}, vol.~8,
  no. 3-4, pp. 279--292, 1992.


\bibitem{Ziebart2008}
B.~D. Ziebart, A.~L. Maas, J.~A. Bagnell, and A.~K. Dey, ``Maximum entropy
  inverse reinforcement learning.'' in \emph{Aaai}, vol.~8.\hskip 1em plus
  0.5em minus 0.4em\relax Chicago, IL, USA, 2008, pp. 1433--1438.


\end{thebibliography}

\end{document}